\documentclass{article}
\usepackage[final]{neurips_data_2023}

\usepackage[utf8]{inputenc} \usepackage[T1]{fontenc}    \usepackage{hyperref}       \usepackage{url}            \usepackage{booktabs}       \usepackage{amsfonts}       \usepackage{nicefrac}       \usepackage{microtype}      \usepackage{xcolor}         

\usepackage{breakcites}
\usepackage{times}
\usepackage{epsfig}
\usepackage{wrapfig}
\usepackage{amsmath, amssymb}
\usepackage{wrapfig}
\usepackage{pdfpages}
\usepackage{tabularx}
\usepackage{arydshln}
\usepackage{multirow}
\usepackage{caption}
\usepackage{subcaption}
\usepackage{hhline}
\usepackage{multirow}
\usepackage{tabularx}
\newcolumntype{Y}{>{\centering\arraybackslash}X}
\usepackage{multirow}
\usepackage{makecell}
\usepackage{mathtools}
\usepackage{stfloats}
\usepackage{cleveref}
\usepackage{appendix}

\graphicspath{{media/}}     \makeatletter
\newcommand{\printfnsymbol}[1]{\textsuperscript{\@fnsymbol{#1}}}
\makeatother

\usepackage{caption,graphicx,newfloat}
\usepackage{fancyvrb}
\usepackage{verbatim}
\usepackage[many]{tcolorbox} 
\DeclareFloatingEnvironment[
  fileext=lob,
  listname={List of Boxes},
  name=Prompt,
  placement=htb!,
]{PROMPT}

\Crefname{PROMPT}{Prompt}{Prompts}
\crefname{PROMPT}{prompt}{prompts}

\newcommand{\parboxc}[1]{\parbox[c]{\hsize}{\vspace{2mm}#1\vspace{2mm}}}
\usepackage{longtable}

\usepackage{todonotes}

 \usepackage{longtable}

\title{ARB: Advanced Reasoning Benchmark for Large Language Models
}

\begin{document}

\newcommand\name{ARB} 
\newcommand\fullname{\textbf{A}dvanced \textbf{R}easoning \textbf{B}enchmark}
\newcommand\email{\texttt{contact@duckai.org}} 

\renewcommand{\thefootnote}{\fnsymbol{footnote}} 
\author{\bf\normalsize{
Tomohiro Sawada$^{1,2,}$ \!\!\! \footnotemark[1] , 
Daniel Paleka$^{1,3}$,
Alexander Havrilla$^{1,2}$, 
Pranav Tadepalli$^{1,2}$,
Paula Vidas$^{1,}$}\\[10pt]
\bf\normalsize{Alexander Kranias$^{1,2}$, 
John J. Nay$^{4,5}$, 
Kshitij Gupta$^{1,6}$,
Aran Komatsuzaki$^{1,2,}$\footnotemark[9] 
}\\[10pt]
\footnotesize{$^1$ DuckAI} 
\footnotesize{$^2$ Georgia Tech} 
\footnotesize{$^3$ ETH Z\"urich} 
\footnotesize{$^4$ Nomos AI} \\
\footnotesize{$^5$ Stanford University Center for Legal Informatics} 
\footnotesize{$^6$ MILA} 
}

\footnotetext[1]{Email: \texttt{tsawada@gatech.edu}.}
\footnotetext[9]{Email: \email{}.}

\maketitle

\begin{abstract}
Large Language Models (LLMs) have demonstrated remarkable performance on various quantitative reasoning and knowledge benchmarks. 
However, many of these benchmarks are losing utility as LLMs get increasingly high scores, despite not yet reaching expert performance in these domains.
We introduce \name{}, a novel benchmark composed of advanced reasoning problems in multiple fields.
\name{} presents a more challenging test than prior benchmarks, featuring problems in mathematics, physics, biology, chemistry, and law. 
As a subset of \name{}, we introduce a challenging set of math and physics problems which require advanced symbolic reasoning and domain knowledge.
We evaluate recent models such as GPT-4 and Claude on \name{} and demonstrate that 
current models score well below 50\% on more demanding tasks. 
In order to improve both automatic and assisted evaluation capabilities, we introduce a rubric-based evaluation approach, allowing GPT-4 to score its own intermediate reasoning steps.
Further, we conduct a human evaluation of the symbolic subset of \name{}, finding promising agreement between annotators and GPT-4 rubric evaluation scores. 
\end{abstract}

\section{Introduction}
\label{sec:introduction}

In recent years, models such as GPT-3 \citep{GPT3}, GPT-4 \citep{openai2023gpt4}, PaLM \citep{PaLM}, and Chinchilla \citep{Chinchilla} have shown increasing performance across a wide variety of natural language tasks ranging from translation to reasoning \citep{bubeck2023sparks, laskar2023systematic}. This rapid progress has been closely tracked and assessed by evaluating LLMs on benchmarks, which test model capabilities on a set of standardized problems.
The GLUE benchmark \citep{wang2019glue} for language understanding was first released in April 2018; 
but models such as BERT \citep{devlin-etal-2019-bert} and GPT-2 \citep{radford2019language} in the following year
were already powerful enough to necessitate the ``SuperGLUE'' benchmark \citep{wang2019superglue}.
Since then, the race between language models and benchmarks has increasingly favored the former. 

Scaling up, model sizes and datasets alike, has led to rapid improvements on various natural language tasks on benchmarks like BIG-bench \citep{srivastava2022beyond} 
and HELM \citep{liang2022holistic}. 
Neural scaling laws \citep{kaplan2020scaling, caballero2023broken, alabdulmohsin2022revisiting} have been used to predict the behavior of large scale models on various metrics. Nevertheless, LLM performance often increases unpredictably \citep{wei2022emergent}, 
especially on tasks that require reasoning abilities.
Predictions of performance on ML benchmarks often underestimate the rate of progress \citep{steinhardt2022oneyearin}. 
Since progress has been faster than anticipated, new benchmarks need to be more difficult.

Models such as ChatGPT have shown the ability to pass  entry-level examinations in fields such as law \citep{bommarito2022gpt}, 
medicine \citep{kung2023performance}, economics \citep{caplan2023gpt}, and mathematics \citep{shakarian2023independent}.
Nevertheless, LLM understanding of many fields is reportedly shallow and unreliable \citep{shapira2023clever}.
\emph{Expert reasoning} in domains with specialized knowledge is essential for automated systems to augment skilled professionals \citep{noy2023experimental}.

In this paper, we introduce a new benchmark dataset, \textbf{\name} (\fullname), 
designed to evaluate expert reasoning abilities in mathematics, physics, chemistry, biology, and law.
To make the benchmark more challenging than previous benchmarks, 
we extract graduate-level tasks from resources intended for domain professionals.
The performance of current models such as GPT-4 on the quantitative parts of {\name} is very low using standard prompting methods.

Our dataset offers improvements over existing benchmarks:
\begin{itemize}
    \vspace{-2pt}
\item Hundreds of problems requiring expert reasoning in quantitative subjects, where LLMs are known to underperform;
    \vspace{-2pt}
\item A large percentage of the problems are short-answer and open response questions, in contrast to the multiple-choice questions that dominated earlier benchmarks.
    \vspace{-2pt}
\end{itemize}

In addition, we propose an automated rubric-based method allowing self-evaluation of intermediate reasoning steps.
While not currently a substitute for human evaluation, rubrics generated by GPT-4 have good coverage, 
and self-evaluation scores track human grading surprisingly well.

We provide the instructions to access the dataset in the supplementary material.

 \section{Related Work}
\label{sec:related}

Improving the reasoning capabilities of LLMs has been a subject of recent interest, with a particular focus on advanced prompting techniques \citep{ChainOfThought, kojima2023large, wang2023selfconsistency, yao2023tree, nye2021work}. Such techniques have seen increasingly successful applications in solving reasoning problems involving commonsense reasoning and mathematics, by promoting active reasoning processes within the LLMs before yielding final answers.

Model architectures such as Minerva \citep{Minerva} have exemplified the enhancement of reasoning capabilities through fine-tuning on extensive datasets covering math and reasoning tasks. This has yielded improved performance across several benchmarks, including MATH \citep{MATH}, GSM8K \citep{GSM8k}, and MMLU \citep{MMMLU}. Concurrently, other lines of research \citep{li2023making, lightman2023lets, GSM8k} have investigated the application of verification techniques to augment and enhance LLM performance.

Most of the aforementioned work has typically evaluated techniques against math benchmarks (e.g., GSM8K \citep{GSM8k}, MATH \citep{MATH}, SVAMP \citep{patel2021nlp}, ASDiv \citep{miao-etal-2020-diverse}, AQuA \citep{ling-etal-2017-program}, MAWPS \citep{koncel-kedziorski-etal-2016-mawps}, MultiArith \citep{roy2016solving}) and commonsense reasoning tasks (e.g., CSQA \citep{CommonSenseQA}, StrategyQA \citep{geva2021did}, HotpotQA \citep{yang2018hotpotqa}). Recently, several new benchmarks have been introduced for reasoning and planning tasks, such as the GPT-Planning Benchmark \citep{valmeekam2023large}, ALERT Reasoning Benchmark \citep{yu2022alert}, 
JEEBench \citep{arora2023llms}),
and  \citep{gendron2023large}. Additionally, comprehensive evaluation suites like the Chain-of-Thought Hub \citep{fu2023chainofthought} have been proposed.

Despite their utility, existing benchmarks 
are limited in difficulty, represent a restricted range of reasoning challenges, 
and do not necessarily mirror real-world tasks demanding complex reasoning. 
Moreover, recent advancements such as Minerva \citep{Minerva} have revealed that these benchmarks may not offer sufficient challenge. 

The rapid progress in LLM capabilities has 
led many to explore using LLMs in the LLM evaluation pipeline.
Apart from using LLMs to generate evaluation tasks \citep{zhang2022dataset-machine-learning, perez2022discovering}, 
LLMs have increasingly been used as a proxy for human evaluation 
\citep{2023arXiv230501937C,2023arXiv230316634L,2023arXiv230204166F,2023arXiv230214520K}. 
Useful LLM-based evaluation for alignment has been done using rubrics \citep{bai2022constitutional}. 
We explore the efficacy of rubrics for evaluation when applied to highly complex math and physics problems.
 \section{Benchmark}
\label{sec:benchmark-description}

The key considerations when building a machine learning benchmark are: 

\begin{itemize}
    \vspace{-2pt}
    \item \textbf{Difficulty.} 
    Most tasks have to be out of reach of current models; a benchmark where many models score over 95\% is not useful for tracking differential AI development.
    \vspace{-2pt}
    \item \textbf{Usefulness.} 
    The tested skills should correlate with generally useful human skills.
    \vspace{-2pt}
    \item \textbf{Ease of evaluation.} 
    It should be straightforward for the model creators to compare the performances of different models. 
    The scores should be interpretable.
    \vspace{-2pt}
    \item \textbf{Minimizing data contamination.}
     A consistent issue with popular benchmarks is that the recent LLMs contain some tasks in their training data \citep{openai2023gpt4}. 
     This leads to overestimation of true model capabilities.
    \vspace{-2pt}
    \item \textbf{Connection to general capabilities.}  
If a model is trained on data similar to the benchmark, it is possible it achieves high performance without generalization or ``intelligence'',
     failing to solve novel tasks of similar difficulty \citep{chollet2019measure}. 
Conversely, problems should not be pathological or overly adversarial, 
     to avoid the dangers of underclaiming 
\citep{bowman2021dangers}.
    \vspace{-2pt}
\end{itemize}

\subsection{Formatting}

The benchmark consists of three types of questions: multiple choice, short answer, and open response, in descending order of proportion in the dataset. 
\begin{itemize}
\item \textbf{Multiple choice} questions consist of a question and four to five possible answers, 
and the correct answer is the one that best answers the question. 
They were sourced from standardized tests, such as the MCAT and bar exam prep, 
and make up a large proportion of the dataset due to their ease of grading. 

\item \textbf{Short answer questions}, on the other hand, ask for final answers in the format of a short phrase or mathematical expression. They were sourced from problem books such as \citet{berkeley}, \citet{putnam}, and physics book series \citet{mechanics}, \citet{electromagnetism}, \citet{quantum}, \citet{optics}, and \citet{statmech}. 
We generally avoided algebraic expressions, because of technical difficulties in the grading process.

A given algebraic expression may have several equivalent forms (e.g. nontrivial functional relations for the functions appearing in the final answer), 
and a grading scheme which accounts for all possible variations across our entire dataset is not feasible. 
Moreover, physics problems often require answers introducing new notation 
that is not explicitly mentioned in the problem statement. 

\item \textbf{Open response} questions are more challenging: they consist of a question and a blank space for the answer. They were sourced from problem books and exams, such as the Harvard PhD comprehensive exams in mathematics \citep{harvard}. 
Such tasks require manual grading. 
These questions are aspirational in nature, as current systems (e.g. ChatGPT) 
cannot produce satisfactory responses, even for the ``elementary''  problems.

\end{itemize}

\begin{table}[h]
\begin{center}
\label{table:types_of_problems}
\caption{Types of problems in the benchmark by subject area.}
\label{tab:types_of_problems_by_subject}
\vspace{4pt}
\begin{tabular}{llc}
Subject  &  Answer Type & Number \\ 
\toprule
\multirow{3}{*}{Mathematics}             
                        & Numerical   & 52 \\ 
                        & Symbolic    & 34 \\ 
                        & Proof-like  & 19 \\ 
\midrule
\multirow{4}{*}{Physics}                 
                        & Numerical               & 80 \\ 
                        & Numerical (w/ image)    & 18 \\ 
                        & Symbolic                & 18 \\ 
                        & Symbolic (w/ image)     & 13 \\ 
\midrule
\multirow{1}{*}{Law}                     
                        & Multiple Choice    & 627 \\ 
\midrule
\multirow{1}{*}{MCAT (Reading)}          
                        & Multiple Choice    & 165 \\ 
\midrule
\multirow{2}{*}{MCAT (Science)}          
                        & Multiple Choice    & 144 \\ 
                        & Multiple Choice (w/ image)   & 37 \\ 
\bottomrule
\end{tabular}
\end{center}
\end{table}

\subsection{Mathematics}

This part of the dataset is the most diverse. It includes contest mathematics problems as well as ``university mathematics'' (i.e. mathematics traditionally taught in universities at the undergraduate and beginning graduate level). The contest problems are sourced from \citet{putnam} and \citet{braymankukush}, and the university mathematics problems are sourced from \citet{berkeley} and \citet{harvard}. The dataset does not include high school contest problems because those are already present in other well-known benchmarks \citep{MATH}. 
The Putnam and Brayman books both contain official solutions, which we also include in the dataset. 
This can be useful for fully automating the grading process, which we leave to future work.

For university mathematics, we pick \citet{berkeley} for its large selection of ``standard'' undergraduate mathematics problems, as well as many problems suitable for the short answer portions. We also select \citet{harvard} because it covers topics that other collections of exams rarely not cover, such as representation theory of finite groups and algebraic topology.

\subsection{Physics}

The physics problems are structured similarly as the math problems.
The main difference is that some physics problems contain figures, 
and there are more problems with numerical answers. 
The problems were sourced from the Major American Universities PhD Qualifying Questions and Solutions series \citep{zhongguo1990major}. 

\subsection{MCAT}
The MCAT test contains multiple choice problems testing 
biology, psychology, chemistry, physics, and reading comprehension. 
The MCAT problems are sampled from the third edition of McGraw-Hill Education 3 MCAT Practice Tests \citep{MCAT} 
and cover both science and reading questions. 
This book was chosen as very few of these problems appear in standard web-searchable sources, 
limiting contamination. 
As in the previous categories, we pick problems which are self-contained.
Because some MCAT science questions are accompanied by images, we accompany such questions with corresponding image files. 

\subsection{Law}

Applying law involves the application logical reasoning, in addition to grasping legal knowledge. 
This makes assessments of legal skills an especially attractive type of language model benchmark, 
where we are attempting to assess the reasoning and intelligence of these models. 
Furthermore, if the models better understand law, they can be more reliable 
and ultimately more useful in real-world applications, 
potentially even increasing the efficiency and transparency of governments more broadly.

Most lawyers in the U.S. go to law school, graduate, then study for the Bar Examination, and then must pass the bar before going on to practice law professionally. To evaluate legal understanding of the models, we use an older Bar Examination practice set that, to the best of our knowledge, is not available online in a way that could have led to its inclusion in training data for the language models that we are assessing. 
The practice bar exam we administer to the various language models covers most major areas of law and therefore it tests legal reasoning and broad U.S. legal knowledge.

 \section{Evaluation}
\label{sec:evaluation}

We evaluate current LLMs on all text-only problems in our dataset. 
Other LLM benchmark papers do not evaluate on multimodal tasks due to the lack of good multimodal models; 
we follow suit.
Given public communications about GPT-4 \citep{openai2023gpt4} and Gemini \citep{palm2},
it is likely the physics and MCAT image problems will be useful for testing multimodal LLMs soon. 

\paragraph{Models} We evaluate ChatGPT (\texttt{gpt3.5-turbo-0301}), GPT 3.5 (\texttt{text-davinci-003}), 
GPT-4 with 8k context length (\texttt{gpt-4-0314}), and Claude (\texttt{claude-v1.3-100k}). 
We evaluate all question types using task-specific instructions and chain of thought. 
In chat models, we put the instructions as the system prompt; otherwise we put them at the beginning of the prompt.

In all problem types, in order to extract the model's final answer, we instruct the model to write its final answer at the end of the response after the delimiter \texttt{ANSWER: }.
We then parse the model generated final answer as the remaining text after the delimiter. 
The response is marked as incorrect if the delimiter is not found.  
Due to the differences in evaluation for multiple choice versus open-ended responses, we adopt multiple evaluation procedures.

\paragraph{Multiple choice}
To evaluate multiple choice questions, we can simply compare the extracted final answer to the ground truth. 
A response is considered correct if the extracted choice matches the ground truth choice. 
With appropriate prompting, all models output a parsable answer > 97\% of the time. 
We conduct a separate manual evaluation on a sampled subset of the questions 
to check that our parsing procedure is not mischaracterizing the true performance of the model.

\paragraph{Numerical}
To evaluate problems with a numerical final answer, we first extract the delimited model answer as above. 
In the physics problems, many answers are in units;
we prompt the model with information about the unit, and instruct it to fully simplify its answer and omit any units.
However, sometimes the model forgets to do either or both, and so we apply a series of regexes to remove units. 
We then attempt to parse the result into a mathematical expression using Python's SymPy library \citep{10.7717/peerj-cs.103}. 
If this parsing fails, the answer is marked as incorrect.
Once parsed, we score a the model answer as correct if $\frac{\left|\texttt{model\_answer} - \texttt{ground\_truth}\right|}{\texttt{ground\_truth}} < 0.01$. 

\paragraph{Symbolic}
Problems with symbolic answers are less structured and harder to parse.
To do so, we again leverage SymPy, first normalizing expressions to contain a default set of variable names and then checking for equivalence up to a permutation of the variables.
However this approach is error-prone and only works for the subset of symbolic responses in a function form.
More advanced responses, such as those containing set notation, require human evaluation. 

\paragraph{Proof-like}
Natural language proofs cannot be evaluated automatically; the authors with training in mathematics grade the proofs.
Further manual human evaluation requires a thorough inspection of the intermediate reasoning steps.
This makes evaluation expensive in practice. 

\paragraph{Model-based evaluation}
To address the difficulties in developing automated metrics for evaluating more advanced problems,
we experiment with two model based approaches.
First, we prompt ChatGPT to grade the equivalence of two symbolic expressions with score options $0$ when the totally incorrect, 
$0.5$ when the symbolic expressions are nearly the same e.g. equivalent up to a constant, and $1$ when they are an exact match.
Our prompting strategy can be found in the supplementary material.

More generally, we evaluate the capabilities of GPT-4 to grade intermediate reasoning chains 
via a \textit{rubric-based} evaluation approach.
For symbolic and proof-like problems, we few-shot prompt GPT-4 to create a 10-point rubric. 
This is done by handwriting a small set of initial rubrics for proof-like problems 
and prompting the model with these examples and the ground truth reference solution. 
The model assigns point values to intermediate steps using the reference solution as a guide.
This process is illustrated in the supplementary material.

With model generated rubrics in hand, we then evaluate each question against its rubric. 
This is done by again prompting GPT-4 to go step by step through the model answer 
and assign partial credit based on the rubric. 
This provides a denser automatic evaluation metric on increasingly unstructured answers.
As a nice side benefit, it makes human evaluation of complex symbolic questions much easier, 
significantly reducing the amount of time required per question.

\subsection{Results}

\begin{figure}[ht]
    \centering
    \includegraphics[scale=0.7]{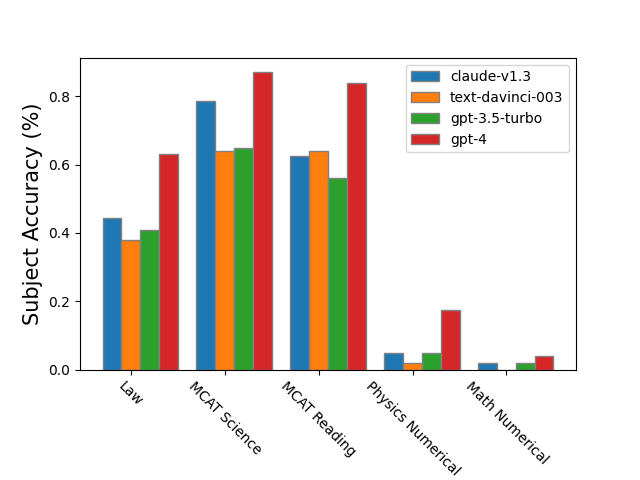}
   \caption{Accuracy of models over automatically scored components of the \name{} benchmark. 
   Numerical questions are evaluated with a relative error threshold of $10^{-2}$.
   }
   \label{fig:len-accs}
\end{figure}

We now evaluate \texttt{gpt-4}, \texttt{gpt-3.5-turbo}, 
\texttt{text-davinci-003}, and \texttt{claude-v1.3} on \name{}. 
The results for the mechanically scored subjects are in \Cref{fig:len-accs}.

We see models generally do quite well on the multiple choice Law and MCAT subsets, 
but struggle significantly on questions with numerical final answers.
GPT-4 is the only model capable of reliably simplifying complex expressions, 
but even GPT-4 struggles to reliably perform arithmetic and symbolic manipulations over long contexts.

On the multiple-choice questions, 
the only model that cannot reliably follow the answer formatting instructions is \texttt{gpt-3.5-turbo}. 
This happens for a variety of reasons, including the model refusing to answer or to commit to a single answer choice.
On the Law benchmark, \texttt{gpt-3.5-turbo} does not output a parsable answer around 25\% of the time.
The other models exhibit this failure in less than 5\% of multiple-choice questions, 
with GPT-4 being correctly parsed over 99\% of the time. 

We see a similarly low performance profile across models on symbolic problems, reported in Table \ref{table:parsed-results}.

\begin{table}[ht]
\centering
\caption{Manually parsed scores for symbolic answer questions.}
\begin{tabular}{@{}lrr@{}}
\\
                       & Math Symbolic     & Physics Symbolic \\
\toprule
\texttt{gpt-4-0314}                  & 18\%    & 28\%      \\
\texttt{gpt-3.5-turbo-0301}          & 12\%    & 6\%      \\
\texttt{text-davinci-003}       & 3\%    & 6\%      \\
\texttt{claude-v1.3-100k}       & 3\%    & 11\%      \\
\bottomrule
\end{tabular}
\label{table:parsed-results}
\end{table}

 \subsection{What Kind of Errors Do LLMs Make?}

The GPT-4 evaluation paper \citep{bubeck2023sparks} classified 
errors GPT-4 makes in single-pass evaluation on GSM8K \citep{GSM8k} and MATH \citep{MATH} into three types:
\emph{arithmetic mistakes}, \emph{misunderstood statement}, and \emph{wrong approach}. 
We make a more fine-grained analysis and extend it to math and physics problems in our dataset.
The results are in \Cref{tab:human-eval-math-problems-gpt-4}. 

The errors current LLMs make on the Mathematics part of \name{} fall into five general types:
\begin{itemize}
    \vspace{-2pt}
    \item {Misunderstanding / answering only a part of the question / misread problem};
    \vspace{-2pt}
    \item {Wrong approach}: the model's early chain of thought does not guess the right approach;
    \vspace{-2pt}
    \item {Logical errors}: the model uses a false implication between two statements;
    \vspace{-2pt}
    \item {Hallucinating facts or theorems}: the model confabulates a statement that is false in general, or not applicable in context;
    \vspace{-2pt}
    \item {Arithmetic/calculation error}: the model multiplies incorrectly, omits a term in an expression, gives a wrong numerical value for a fraction, and other similar mistakes.
    \vspace{-2pt}
\end{itemize}

We grade GPT-4 using the above as a guideline.
Our grading of the model's CoT answers is not mutually exclusive; 
if the model both uses an approach that doesn't go anywhere and makes a calculation error in it,  we count it towards both categories.
Note that the errors might not be independent: arithmetic mistakes could be more or less frequent in wrong approach solutions as opposed to the solutions with correct idea. We notice that the model is likely to make incorrect simplifications to get to some final answer in approaches that cannot work; this is expected, as prompting the model to produce a solution with a final answer often leads it to produce \emph{some} final answer by any means.

When the model outputs a chain of implications, 
it is not always clear whether some false statement is due to a logical error, 
or it is a straight-out confabulation.  We merge those two error types in \Cref{tab:human-eval-math-problems-gpt-4}.

\begin{table}[t]
\centering
\caption{Mistakes on mathematics and physics problems in \name{}, GPT-4.}
\vspace{4pt}
\begin{tabular}{@{}lrrrrrr@{}}
 & Misread             & Wrong       & Logical error                 & Arithmetic  &  Correct  & Correct \\
 & problem             &  approach   & or hallucination              & mistake      &  answer   & reasoning \\
\toprule
Math Numerical & 0\% & 25\% & 88\% & 48\% & 3\%  & 3\% \\
Math Symbolic & 16\% & 50\% & 29\% & 4\% & 16\%  & 16\% \\
Math Proof-like & 5\% & 50\% & 72\% & 16\% & n/a & 5\% \\ Physics Numerical & 0\% & 80\% & 53\% & 6\% & 6\%  & 6\% \\
Physics Symbolic & 0\% & 37\% & 68\% & 31\% & 28\%  & 12\% \\
\bottomrule
\end{tabular}
\label{tab:human-eval-math-problems-gpt-4}
\end{table}

Some problems ask for multiple things to be proven or calculated. 
Our graders gave the model a score of 0.5 if it correctly derived at least half of the "subproblems" (for example, homology groups of a given manifold). 
With this more benevolent form of grading, the performance of GPT-4 on the Proof-like problems jumps to 16\%.
Where applicable, slight discrepancy with automatic evaluation is also possible due to the error tolerance.

We note that many of the problems in Physics Symbolic have correct symbolic answers even when there are flaws in the chain of thought reasoning of GPT-4. 
This is likely due to some kind of memorization, although not necessarily from the same sources: see Table \ref{table:memorization} for an example.

It is possible that our graders underestimate the rate of arithmetic mistakes in some cases, 
especially when the approach is clearly wrong, or it is not clear whether a given error is due to faulty reasoning or due to a missed term in the calculations.

For the larger subsets (see \Cref{tab:types_of_problems_by_subject}), 
we subsample the problems to between 20 and 40 per subject area; 
this is enough for a ballpark estimate of the frequency of different errors,
and is not worth increasing because attributing error types is inherently fuzzy.

 \section{Model-based Rubric Evaluation}
\label{sec:rubric-eval}

As reasoning tasks increase in complexity, it gets harder to evaluate model performance.
Symbolic final answers are in some cases difficult to grade automatically.
Further, we are often more interested in the correctness of the 
reasoning used to produce the final answer; 
but evaluating intermediate reasoning steps requires expert human supervision. 
An ideal solution would be to use LLMs as evaluators based on a reference solution; 
unfortunately, there are major reliability issues.

To improve reliability, we propose generating \emph{rubrics} 
as an important component of the evaluation process.
The model generates the rubric from the reference solution, 
then evaluates any solution based on the generated rubric.
To aid rubric generation, we give few-shot examples of human-written rubrics to the rubric-generating model run.
We study this approach by conducting a human evaluation of GPT-4 generated rubrics 
and the GPT-4 grading of its own solutions using the generated rubrics.

We rate the quality of GPT-4 generated rubrics by hand in the first two rows of Table \ref{table:rubric-quality}.
Likert scores from 1-5 are assigned to both the \textit{coverage} of the rubric, 
i.e. how well it captures key subproblems, and the point breakdown. 
Rubric quality scores are reported in Table \ref{table:rubric-results} 
for symbolic and proof-like problems.  
We find GPT-4 designs rubrics which cover the crucial solution steps well,
but struggles to properly allocate points to each step based on relative importance.
However, it is much better than GPT-3.5-turbo, 
which tends to over-allocate points to only one or two solution steps.

\begin{table}[ht]
\begin{center}
\caption{Evaluations of rubric quality and GPT-4 rubric evaluation failure cases. 
Rubric coverage and rubric point spread are on a 1-5 Likert scale. 
Alternative solutions is the percentage of correct solutions found not covered by the rubric.
Extra/reduced credit track how often GPT-4 erroneously assigns or deducts points.
Hallucinated rubric tracks how often GPT-4 assigns points by referring to a rubric item 
not actually present in the rubric.
}
\label{table:rubric-quality}
\begin{tabular}{@{}lrrr@{}}
\\
    & Physics Symbolic & Math Symbolic  & Proof-like \\
 \toprule
 Rubric coverage & 4.42 & 4.26 & 3.94 \\
 Rubric point spread & 4.16 & 4.00 & 4.06 \\
Alternative solutions & 5\% & 2\% & 0\% \\
 Extra credit & 27\% & 18\% & 40\% \\
 Reduced credit & 11\% & 12\% & 5\% \\
 Hallucinated rubric & 0\% & 15\% & 0\% \\
 \bottomrule
\end{tabular}
\end{center}
\end{table}

The obvious limitation of rubric scoring is the case of correct solutions not covered by the rubric.
We find that on our benchmark, GPT-4 rarely generates a fully or even mostly partially correct solution 
that does not follow the rubric.
Once done rating the model generated rubrics, 
we then manually grade GPT-4's solutions according to each rubric 
and compare the result to GPT-4's evaluation. 
We also annotate, for each problem, both whether GPT-4 assigns credit inappropriately 
or fails to assign credit when it should. 

\begin{table}[ht]
\begin{center}
\caption{Average scores (out of 10 points) when assigned by human annotators versus GPT-4. 
Correlation is the Pearson correlation coefficient between the two scores, over all problems.
}
\label{table:rubric-results}
\begin{tabular}{@{}lrrr@{}}
\\
    & Physics Symbolic & Math Symbolic  & Proof-like \\
 \toprule
 Human eval score & 5.00 & 3.13 & 2.65 \\
 Model eval score & 5.05 & 3.37 & 3.8 \\
 Correlation & 0.91 & 0.78 & 0.82 \\
 \bottomrule
\end{tabular}
\end{center}
\end{table}

We find a moderately high correlation between GPT-4's evaluation score and the manual score. 
In some cases, the model, assigns an extra point or two when compared to the annotated rubric score. 
However, the self-eval score almost never deviates more than two points from the ground truth.
The main failure mode we detect is the assignment of partial credit to attempted solutions 
completely outside the problem rubric, where the human evaluation score is always zero.
Taken together, we believe these results suggest that
rubric-based evaluation is a promising automated evaluation method.

Having established rubric-based evaluation as a (imperfect) proxy for correctness,
we now comment on the GPT-4 performance graded by the rubric.
Table \ref{table:rubric-results} shows GPT-4 is
best at generating correct intermediate reasoning steps for physics questions.
Inspecting the model outputs suggests that GPT-4 is good 
at recalling relevant and useful concepts in physics for solving the relevant problem; 
however, it can struggle with the mathematical manipulations required to solve the problem.
The model is worse at recognizing the correct concepts 
and formulating an appropriate plan for the math questions, 
particularly for proof-like problems.
 \section{Limitations and Conclusion}
\label{sec:conclusion}
In this paper, we presented \name{}, a novel benchmark for evaluating advanced reasoning capabilities in large language models. Our dataset is composed of various problems from the sciences and law, sourced from graduate-level exams and professional resources.  Despite advancements in current LLMs, their performance remains very low on the quantitative subjects,
in \name's tasks. 
We also proposed a rubric-based self-evaluation method, enabling LLMs to grade their own reasoning. 
This method is not yet reliable enough to replace human grading. 
We hope this method can be extended to more reliable and cheap testing of complex model outputs.

As with all other benchmarks that are not created anew and kept secret, it is possible there is data contamination.
For example, the MCAT books are not available for free in most jurisdictions,
but it certainly possible that some model creators have trained on it anyway.

Finally, the benchmark does not remotely cover all aspects of human ability; a model solving this benchmark perfectly could still be much worse than most educated people in many aspects. Nevertheless, we hope that increasing the difficulty standards helps the research community ground the performance of increasingly powerful models more accurately.

\begin{ack}
\label{acknowledgements}
We thank Jeffrey Deng for developing and documenting the API, and building the project website. 
We would also like to thank Raunak Chowdhuri for helpful comments, and Zhangir Azerbayev for useful discussions early on in the project. TS is supported by NSF grant 1745583. 
\end{ack} 

{
\footnotesize
\bibliographystyle{plainnat}
\bibliography{refs}
}

 \begin{appendices} 
\section{Datasheet}
We present the data card, following the format proposed by \citet{2022arXiv220401075P}.

\textbf{Dataset Owners.} \email{}.

\begin{table}[h]
\begin{center}
\caption{Data overview.}
\label{table:data_overview}
\vspace{2pt}
\begin{tabular}{llc}
Subject  &  Task Type & Source \\ 
\toprule
\multirow{3}{*}{Mathematics}             
                        & Contest problems   & \citet{putnam, braymankukush} \\ 
                        & University math proof  & \citet{berkeley, harvard} \\ 
\midrule
\multirow{1}{*}{Physics}                 
                        & PhD qualifying exam               & \citet{zhongguo1990major} \\
\midrule
\multirow{1}{*}{Law}                     
                        &  US Law Standardized Exam   &  \citet{barbri2007} \\ 
\midrule
\multirow{1}{*}{MCAT}          
                        & Reading comprehension   & \citep{MCAT} \\ 
\midrule
\multirow{1}{*}{MCAT}          
                        & College science    & \citep{MCAT} \\ 
\bottomrule
\end{tabular}
\end{center}
\end{table}

\textbf{Dataset Overview.} See Table \ref{table:data_overview}.

\textbf{Risk and Mitigation.} There is little risk associated with this dataset, as it is intended for benchmarking reasoning capabilities of models, and it is too small to be used for advancing capabilities.

\textbf{Maintenance Status.} 
Limited Maintenance. The data will not get major updates, but any technical issues will be
addressed.

\textbf{Maintenance Plan.} Any technical issues will be addressed. 
\begin{itemize}
    \item \textbf{Versioning.} No new versions are planned.
    \item \textbf{Updates.} Updates are limited to bug and error fixes.
    \item \textbf{Errors.} Error handling will be considered case by case.
    \item \textbf{Feedback.} \email{}.
\end{itemize}

\textbf{Example: Typical Data Point.} Each data point of the dataset consist of a pair of problem statement and ground truth solution. \Cref{table:rubric-example}, \Cref{table:mcat} and \Cref{table:math} include problem statement and ground truth solution of typical data points. 

\textbf{Sensitive Human Attributes.} We have not found any sensitive human attributes in our dataset.

\textbf{Data Distributions.} 
Table \Cref{table:types_of_problems} shows the number of problems for each subject area and answer type. Text entries (problem statement, ground truth solution, ground truth answer) for all categories are in LaTeX (although obviously, the non-quantitative subjects have very few mathematical expressions).   \section{Dataset format}

The benchmark dataset is available in .jsonl format, containing problem statements, ground truth solutions, 
and final ground truth answers for each entry. 
We additionally include metadata such as subject names and problem topics, where available.

We chose the four subject areas discussed earlier for several reasons.
Primarily, the dataset focuses on math and physics, as these subjects present more challenging problems than existing benchmarks. 
However, to ensure a comprehensive evaluation of models, we also included subjects like Law and MCAT. 
This inclusion allows for assessing model performance across a wider range of technical domains, beyond the quantitative sciences.

Although previous works have evaluated recent models on law \citep{katz2023gpt},
we draw upon the established importance of broad benchmarks like SuperGLUE \citep{wang2019superglue}. 
Making a benchmark more comprehensive expands the evaluation scope while enhancing the dataset's significance in the wider AI research context.

\section{Accessing the dataset}
\label{app:accessing}

We facilitate access to the dataset through API calls\footnote{The API can be accessed through the following: \href{https://arb.duckai.org/api/lib}{https://arb.duckai.org/api/lib}. The documentation can be found here: \href{https://app.swaggerhub.com/apis-docs/arb-dataset/arb-api/1.0.5}{https://app.swaggerhub.com/apis-docs/arb-dataset/arb-api/1.0.5}}. 
The problems can be accessed by the different splits and subject areas. 

We use an API to host the data (rather than making it publicly available on popular platforms such as HuggingFace or Github) out of concern for data contamination. 
Many models released in the last few years have been trained on benchmark datasets \cite{openai2023gpt4}, an approach that could inadvertently inflate the model's performance on the associated benchmark. 
There are several proposed strategies to mitigate this problem, including dataset poisoning and canary text methods.However, restricting dataset accessibility to web crawlers seemed to be the only way to ensure integrity of our dataset for future uses. 

 \begin{table}[ht]
    \centering
    \caption{Example model generated rubric and evaluation. Evaluation with a rubric reveals the model begins with the model makes good partial progress but is unable to arrive at the correct final answer.}
    \label{table:rubric-example}
    \vspace{3mm}
    \noindent
    \begin{tabularx}{\textwidth}{p{0.11\textwidth}p{0.89\textwidth}}
    \toprule
    \parboxc{Problem Statement} & \parboxc{Find the maximum area of all triangles that can be inscribed in an ellipse with semiaxes $a$ and $b$, and describe the triangles that have maximum area.} \\
    \midrule
    \parboxc{Ground Truth Solution} & \parboxc{Using the parameterization
Using the parameterization

$$
x=a \cos t, y=b \sin t,
$$

a triple of points on the ellipse is given by

$$
\left(a \cos t_{i}, b \sin t_{i}\right), \quad i=1,2,3 .
$$

So the area of an inscribed triangle is given by

$$
\frac{1}{2}\left|\begin{array}{lll}
1 & a \cos t_{1} & b \sin t_{1} \\
1 & a \cos t_{2} & b \sin t_{2} \\
1 & a \cos t_{3} & b \sin t_{3}
\end{array}\right|=\frac{a b}{2}\left|\begin{array}{lll}
1 & \cos t_{1} & \sin t_{1} \\
1 & \cos t_{2} & \sin t_{2} \\
1 & \cos t_{3} & \sin t_{3}
\end{array}\right|
$$

which is $a b$ times the area of a triangle inscribed in the unit circle. In the case of the circle, among all inscribed triangles with a given base $2 w(0<w \leqslant 1)$, the one of maximum area is an isosceles triangle whose area equals

$$
g(w)=w\left(1+\sqrt{1-w^{2}}\right) .
$$

Using elementary calculus one finds that the maximum of $g$ on the interval $0 \leqslant w \leqslant 1$ occurs at $w=\sqrt{3} / 2$, corresponding to an equilateral triangle, and equals $3 \sqrt{3} / 4$. Alternatively, fixing one side of the triangle as the basis, we easily see that among all the inscribed triangles the one with the greatest area is isosceles because of the maximum height, showing that the angle at the basis is the same. Fixing another side we see that the triangle is indeed equilateral. Hence, the area is maximal when

$$
t_{2}=t_{1}+\frac{2 \pi}{3} \text { and } t_{3}=t_{2}+\frac{2 \pi}{3}
$$

that is, when the corresponding triangle inscribed in the unit circle is regular.

For the ellipse with semiaxes $a, b$, this corresponds to an inscribed triangle with maximum area equals $3 a b \sqrt{3} / 4$. }
 \\
 \midrule
\parboxc{Model Generated Rubric} & \parboxc{
\begin{itemize}
\item Parameterize the ellipse using $x=a \cos t, y=b \sin t$ (2 points).
\item Write the expression for the area of an inscribed triangle using the determinant formula (2 points):

\begin{equation}
    \frac{ab}{2}\left|\begin{array}{lll}
    1 & \cos t_{1} & \sin t_{1} \\
    1 & \cos t_{2} & \sin t_{2} \\
    1 & \cos t_{3} & \sin t_{3}
    \end{array}\right|
\end{equation}

\item Connect the problem to the maximum area of an inscribed triangle in the unit circle (2 points).
\item Correctly identify that the maximum area occurs when the inscribed triangle is equilateral (2 points).
\item Find the maximum area of an inscribed triangle in the ellipse, and provide the correct answer (2 points):

\begin{equation}
    \frac{3ab\sqrt{3}}{4}
\end{equation}
\end{itemize} 
} \\
\bottomrule
\end{tabularx}
\end{table}

\begin{table}[ht]
\centering
\caption*{\Cref{table:rubric-example} (continued)}
\begin{tabularx}{\textwidth}{p{0.11\textwidth}p{0.89\textwidth}}
\toprule
\parboxc{Model\newline Response} & \parboxc{To maximize the area of the inscribed triangle, we can use the following approach:

1. Parameterize the ellipse: The equation of an ellipse with semiaxes $a$ and $b$ is given by $\frac{x^2}{a^2} + \frac{y^2}{b^2} = 1$. We can parameterize the ellipse using the parameter $t$ as follows: $x = a\cos(t)$ and $y = b\sin(t)$, where $0 \le t \le 2\pi$.

2. Determine the vertices of the triangle: Let $A(a\cos(t_1), b\sin(t_1))$, $B(a\cos(t_2), b\sin(t_2))$, and $C(a\cos(t_3), b\sin(t_3))$ be the vertices of the inscribed triangle, where $0 \le t_1, t_2, t_3 \le 2\pi$.

3. Calculate the area of the triangle: Using the Shoelace Theorem, the area of the triangle is given by:

\begin{align*}
    \text{Area} = \frac{1}{2} \left| \sum_{i=1}^3 x_i y_{i+1} - \sum_{i=1}^3 x_{i+1} y_i \right|
\end{align*}

where $(x_4, y_4) = (x_1, y_1)$.

Substituting the parameterized coordinates, we get:

\begin{align*}
    \text{Area} = \frac{1}{2} | a^2 \cos(t_1)\sin(t_2) + a^2 \cos(t_2)\sin(t_3) \\ + a^2 \cos(t_3)\sin(t_1) - a^2 \cos(t_2)\sin(t_1) - a^2 \cos(t_3)\sin(t_2) - a^2 \cos(t_1)\sin(t_3) |
\end{align*}

Simplifying the expression, we get:

\begin{align*}
    \text{Area} = \frac{a^2}{2} | \cos(t_1)\sin(t_2) - \cos(t_2)\sin(t_1) +\\ \cos(t_2)\sin(t_3) - \cos(t_3)\sin(t_2) + \cos(t_3)\sin(t_1) - \cos(t_1)\sin(t_3) |
\end{align*}

Using the sine difference identity, $\sin(A - B) = \sin(A)\cos(B) - \cos(A)\sin(B)$, we can rewrite the area as:

$$\text{Area} = \frac{a^2}{2} \left| \sin(t_1 - t_2) + \sin(t_2 - t_3) + \sin(t_3 - t_1) \right|$$

4. Maximize the area: Notice that the maximum value of each sine term is 1, and this occurs when the angles $t_1, t_2, t_3$ are equally spaced around the ellipse, i.e., when the triangle is equilateral. In this case, the area becomes:

$$\text{Area} = \frac{a^2}{2} \left| 1 + 1 + 1 \right| = \frac{3a^2}{2}$$

5. Relate the maximum area to the semiaxes: Since the triangle is equilateral, the ratio of the area of the triangle to the area of the ellipse is constant. The area of the ellipse is given by $\pi ab$. Thus, we can write:

$$\frac{\text{Area}_{\text{max}}}{\pi ab} = \frac{\frac{3a^2}{2}}{\pi ab}$$

Solving for the maximum area, we get:

$$\text{Area}_{\text{max}} = \frac{3a^2}{2} \cdot \pi ab = \frac{3\pi a^3 b}{2}$$

Thus, the maximum area of an inscribed triangle in an ellipse with semiaxes $a$ and $b$ is $\frac{3\pi a^3 b}{2}$, and the triangles with maximum area are equilateral.

ANSWER: $\frac{3\pi a^3 b}{2}$ } \\
\bottomrule
\end{tabularx}
\end{table}

\begin{table}[ht]
\centering
\caption*{\Cref{table:rubric-example} (continued)}
\begin{tabularx}{\textwidth}{p{0.11\textwidth}p{0.89\textwidth}}
\toprule
\parboxc{Rubric Based Evaluation} & \parboxc{
1. The student parameterized the ellipse correctly using $x=a\cos(t)$ and $y=b\sin(t)$. (2 points)

2. The student used the Shoelace Theorem to find the area of the triangle, but did not use the determinant formula as specified in the rubric. (0 points)

3. The student did not explicitly connect the problem to the maximum area of an inscribed triangle in the unit circle. (0 points)

4. The student correctly identified that the maximum area occurs when the inscribed triangle is equilateral. (2 points)

5. The student found the maximum area of the inscribed triangle in the ellipse, but their answer is incorrect. The correct answer is $\frac{3ab\sqrt{3}}{4}$, not $\frac{3\pi a^3 b}{2}$. (0 points)

Rubric Score: 4 points } \\
\bottomrule
\end{tabularx}
\end{table}

\section{Model-based Rubric Eval}

\Cref{table:rubric-example} presents a GPT-4 generated rubric and self-evaluation for symbolic math questions.
GPT-4 does a good job breaking the problem into important sub-parts of the reference solution. 
Further, it ensures a well-balanced point distribution among sub-problems, assigning similar point values to each sub-part.
In contrast, less powerful models like GPT-3.5-turbo struggle to generate good rubrics,
as they tend to under-specify sub-problems and allocate too many points to irrelevant steps.

The main downside on evaluating solutions only on the final answer is that the metric is not smooth;
a small mistake in the middle of the solution always leads to an incorrect final answer.
Grading using a rubric shows that GPT-4 generates some useful ideas, 
for example the correct formula for parameterizing an ellipse. 
The model also correctly identifies that the question's area is optimized by an isosceles triangle.
Despite this, it is unable to correctly compute the final answer due to an earlier mistake in the response.
This indicates that GPT-4 has some problem-solving abilities, 
but struggles to detect or recover from earlier errors in generation.

\subsection{Using ChatGPT for Symbolic Evaluation}

Unlike GPT-4, GPT-3.5-turbo is not to write rubrics with good coverage of the reference solution and a fair point breakdown. 
Often the model will over-simplify the rubric and allocate far too many points to non-essential parts of the problem. 
However, GPT-3.5-turbo does possess some ability to reason about complex symbolic expressions. Motivated by this, we asked the model to grade the final answers to symbolic math and physics problems. While much easier to grade than intermediate reasoning steps, more involved symbolic expressions still require human evaluation to compare accurately. Using cheap models like GPT-3.5-turbo to automate this symbolic equivalence boosts our abilities to evaluate models on this more complicated class of reasoning problems. We prompt GPT-3.5-turbo to compare extracted model generated symbolic final answers from GPT-4 to the reference answer and record results in Table \ref{table:symb-eval}.

\begin{table}[ht]
\begin{center}
\caption{Performance of GPT-3.5-turbo on symbolic equivalence versus human ground truth. 
The model achieves a false positive rate of 0, but allows many false negatives. 
This suggests there are many cases in which the model does not recognize equivalent expressions.
}
\begin{tabular}{@{}lrr@{}}
\\
    & Physics Symbolic & Math Symbolic \\
 \hline
 True positives & 3 & 4 \\
 False positives & 0 & 0 \\
 True negatives & 9 & 22 \\
 False negatives & 6 & 8 \\
 Accuracy & 0.67 & 0.76 \\
 \hline
\end{tabular}
\label{table:symb-eval}
\end{center}
\end{table}

GPT-3.5-turbo is surprisingly good at not equating non-equivalent symbolic statements, achieving a false positive rate of 0. 
However, this comes at of the model often not assigning credit when it should, leading to an underestimate of the true performance of the graded answers on symbolic tasks. In particular, GPT-3.5-turbo often fails when comparing expressions with more than 3 variables. Furthermore, we observe in several cases the student model GPT-4 is able to correctly generate the final answer, but does not present it in the expected format, making it impossible for GPT-3.5-turbo to correctly compare against the ground truth reference. These errors could potentially be avoided by giving the model access to the entire student generated reasoning trace, but we expect this to be a minimal help and perhaps even harm performance.

\section{Interesting Examples}

\begin{table}[ht]
\centering
\caption{Problem, solution and GPT-4 response for a question from the MCAT part of the benchmark. The model mistakes two copies of a DNA with the two amino acid strands in a DNA.}
\vspace{3mm}
\label{table:mcat}
\noindent
\begin{tabularx}{\textwidth}{p{0.12\textwidth}p{0.88\textwidth}}
\toprule
\parboxc{Problem \\ Statement} & \parboxc{If you replaced the thymine in a double helix with radioactive thymine and you allowed DNA replication to occur once, which of the following results would you expect? Choices: \newline A: 
The DNA would not be radioactive in either of the double helices.\newline 
B: The mRNA made from the double helices would be radioactive.\newline 
C: The DNA in each of the double helices would be radioactive.\newline 
D: The DNA in one of the double helices would be radioactive but not in the other one.} \\
\midrule
\parboxc{Ground Truth \\ Solution} & \parboxc{The correct answer is C. During semiconservative replication, the DNA helix unwinds so that both strands of DNA can serve as a template. Both strands are copied, producing two double helices. Each one consists of a template strand and a newly synthesized strand.} \\
\midrule
\parboxc{Model \\ Response} & \parboxc{First, let's consider the process of DNA replication. DNA replication is a semi-conservative process, which means that each new double helix formed contains one original strand and one newly synthesized strand. In this case, the original strand contains radioactive thymine.\newline

Now let's analyze the answer choices:\newline

A: The DNA would not be radioactive in either of the double helices.\newline
This choice is incorrect because we know that the original strand contains radioactive thymine, and since replication is semi-conservative, each new double helix will have one original strand with radioactive thymine.\newline

B: The mRNA made from the double helices would be radioactive.\newline
This choice is unrelated to the question since it concerns mRNA instead of the DNA replication process. Therefore, we can rule out this choice.\newline

C: The DNA in each of the double helices would be radioactive.\newline
This choice suggests that both strands in each new double helix would be radioactive. However, we know that only the original strand contains radioactive thymine, and the newly synthesized strand does not. Thus, this choice is not correct.\newline

D: The DNA in one of the double helices would be radioactive but not in the other one.\newline
This choice aligns with the concept of semi-conservative replication. Each new double helix contains one original strand (with radioactive thymine) and one newly synthesized strand (without radioactive thymine).\newline

Therefore, the correct answer is:\newline

ANSWER: D} \\
\bottomrule
\end{tabularx}
\end{table}

\begin{table}[ht]
\centering
\caption{Problem, solution and GPT-4 sesponse for a question from the Math Numerical part of the benchmark. 
The model often returns a numerical answer after doing some internal black-box numerical calculation, which is rarely correct.}
\vspace{2mm}
\label{table:math}
\noindent
\begin{tabularx}{\textwidth}{p{0.11\textwidth}p{0.89\textwidth}}
\toprule
\parboxc{Problem Statement} & \parboxc{Compute $\displaystyle\int_{0}^{\pi} \frac{x \sin x}{1+\sin ^{2} x} d x .$} \\
\midrule
\parboxc{Ground Truth Solution} & \parboxc{We use the example from the introduction for the particular function $f(x)=\frac{x}{1+x^{2}}$ to transform the integral into$$ \pi \int_{0}^{\frac{\pi}{2}} \frac{\sin x}{1+\sin ^{2} x} d x . $$This is the same as$$ \pi \int_{0}^{\frac{\pi}{2}}-\frac{d(\cos x)}{2-\cos ^{2} x}, $$which with the substitution $t=\cos x$ becomes$$ \pi \int_{0}^{1} \frac{1}{2-t^{2}} d t=\left.\frac{\pi}{2 \sqrt{2}} \ln \frac{\sqrt{2}+t}{\sqrt{2}-t}\right|_{0} ^{1}=\frac{\pi}{2 \sqrt{2}} \ln \frac{\sqrt{2}+1}{\sqrt{2}-1}$$ } \\
\midrule
\parboxc{Model \\ Response} & \parboxc{To compute the integral $$ \int_{0}^{\pi} \frac{x \sin x}{1+\sin ^{2} x} d x , $$ we can use integration by parts. Let $u = x$ and $\displaystyle dv = \frac{\sin x}{1+\sin^2 x} dx$. Then, we have $du = dx$ and $\displaystyle v = \int \frac{\sin x}{1+\sin^2 x} dx$.
To find the integral of $v$, we can use the substitution method. Let $t = \sin x$, then $dt = \cos x dx$. The integral becomes:\newline $$\int \frac{t}{1+t^2} dt = \frac{1}{2} \int \frac{2t}{1+t^2} dt = \frac{1}{2} \ln(1+t^2) + C$$\newline Now, we can substitute back $t=\sin x$, and we get:\newline $$v = \frac{1}{2} \ln(1+\sin^2 x) + C$$\newline Now, we can apply integration by parts:\newline$$\int_{0}^{\pi} \frac{x \sin x}{1+\sin ^{2} x} d x = uv\big|_0^\pi - \int_{0}^{\pi} v du = \left[x \cdot \frac{1}{2} \ln(1+\sin^2 x) \right]_0^\pi - \int_{0}^{\pi} \frac{1}{2} \mathrm{ln}(1+\sin^2 x) dx$$\newline  The first term is:\newline $$\left[x \cdot \frac{1}{2} \ln(1+\sin^2 x) \right]_0^\pi = \pi \cdot \frac{1}{2} \ln(1) - 0 \cdot \frac{1}{2} \ln(1+\sin^2 0) = 0$$\newline Now, we need to compute the second term:\newline $$-\int_{0}^{\pi} \frac{1}{2} \ln(1+\sin^2 x) dx$$\newline Unfortunately, this integral does not have a closed-form solution in terms of elementary functions. However, it is possible to compute its numerical value using numerical integration methods, such as the trapezoidal rule, Simpson's rule, or Gaussian quadrature.\newline Using numerical integration, we can approximate the value of the integral to be approximately $-1.9348$. Therefore, the final answer is: ANSWER: $-1.9348$. } \\
\bottomrule
\end{tabularx}
\end{table}

\paragraph{Math Example.}

One common way GPT-4 gets numerical questions wrong is when it do some formal manipulations follows by a blackbox numerical calculation after which the model spits out an (incorrect) answer. (Table \ref{table:math}) The formal manipulation does not simplify the problem, and so all the work of solving the problem is done in the final step of the calculation.

\paragraph{MCAT Example.}

GPT-4 get's confused when meanings of words are implicit in prerequisite knowledge or contexts. In one example about DNA replication (Table \ref{table:mcat}), the model correctly identifies that the radioactive thymine is present in the two strands of nucleotides from the original DNA, it fails to deduce that both of the resulting double helices are radioactive. This seems to be because the model confuses the word "DNA" with "strands" of the DNA. When looking at choice C, the model (incorrectly) assumes that each of the four strands in the new double helices are radioactive, when it is clear from context that the choice is referring to the radioactive molecule being present somewhere in each double helix (not necessarily in each strand). Because of this misconception, the model chooses D. 

\paragraph{Law Example.}

An unexpected mistake from GPT-4 in answering law questions is where the model reads too much into an answer choice. For example, GPT-4 incorrectly produced this answer and reasoning:

\begin{quote}
``B: This answer choice suggests that when two crossing offers are identical, one will be treated as an offer and the other as an acceptance. This accurately reflects the situation between Mom and Huck, as they both agreed on the same price.''
\end{quote}

And made this final answer choice:
\begin{quote}
``Yes, because when two crossing offers are identical in import, one will be treated as an offer and the other as an acceptance.''
\end{quote}

The error GPT-4 made is treating the statement in the answer choice (``when two crossing offers are identical, one will be treated as an offer and the other as an acceptance'') as a fact, and then making a choice based on that fact being true.  Better prompting could likely precent errors of this type.

\begin{table}[ht]
\centering
\caption{Problem, solution and GPT-4 response for a question from the law part of the benchmark.}
\vspace{2mm}
\label{table:law}
\noindent
\begin{tabularx}{\textwidth}{p{0.11\textwidth}p{0.89\textwidth}}
\toprule
\parboxc{Problem Statement} & \parboxc{Mom, a wealthy woman, wished to buy her son, Sylvester, an expensive Rolls-Royce for a wedding present. She visited Dealership, a Rolls-Royce dealer, several times, looked at many cars, and discussed possible sales prices with Huck, a salesman. On May 15, after much discussion, Huck and Mom signed a writing that stated, "If we can agree on price on or before June 1, Mom agrees to buy and Huck, for Dealership, agrees to sell one yellow Rolls-Royce Silver Streak, serial number XO857623." On May 20, Mom dispatched a letter to Huck stating, "I will buy the Rolls-Royce for $ 150,000$." On the same day, Huck dispatched a letter to Mom, stating, "I will sell the Rolls-Royce for $ 150,000$." Has a valid contract been formed between Dealership and Mom?\\

Answer Candidates:\\
A.) Yes, because the May 15 writing constitutes a contract with a missing price term, and that term was filled by the crossing offers.\\
B.) Yes, because when two crossing offers are identical in import, one will be treated as an offer and the other as an acceptance.\\
C.) No, because there were two crossing offers and no acceptance; hence there was no mutual assent.\\
D.) No, but the result would be different if Mom were a merchant.} \\
\midrule
\parboxc{Ground Truth Solution} & \parboxc{C} \\
\midrule
\parboxc{Model \\ Response} & \parboxc{First, let's analyze the May 15 writing. The writing states that if the price can be agreed upon by June 1, Mom and Huck will enter into a contract for the purchase of the specific Rolls-Royce. At this point, there is no agreement for the price, so there is no valid contract formed yet.

Next, let's examine the crossing offers on May 20. Mom sent a letter stating she would buy the car for \$150,000, and Huck sent a letter stating he would sell the car for \$150,000. Both letters were dispatched on the same day, and they agreed on the same price. Since the price was one of the main terms that needed to be agreed upon, and both parties agreed on the same price, this could be seen as one party making an offer and the other accepting. Thus, a valid contract could be formed.

Now, let's look at the answer choices:

A: This answer choice argues that the May 15 writing is a contract with a missing price term. However, the May 15 writing itself stated that an agreement would only be formed if the price was agreed upon. Therefore, this answer choice is incorrect.

B: This answer choice suggests that when two crossing offers are identical, one will be treated as an offer and the other as an acceptance. This accurately reflects the situation between Mom and Huck, as they both agreed on the same price.

C: This answer choice argues that there was no mutual assent because there were two crossing offers and no acceptance. However, since both offers agreed on the same price, one could be seen as an offer and the other as an acceptance, forming a valid contract.

D: This answer choice states that the result would be different if Mom were a merchant. However, the fact that Mom is not a merchant does not affect the formation of a valid contract in this situation.

Given the analysis of the answer choices, the correct answer is:

ANSWER: B} \\
\bottomrule
\end{tabularx}
\end{table}  \section{Memorization}

While it is difficult to quantify potential problem leakage and memorization of similar problems, 
some outputs suggest this might be a relevant factor.
In Table \ref{table:memorization}, GPT-3.5-turbo appears to be memorizing the answer to this symbolic physics problem, 
given that it is getting the formula for capacitance $C$ correct despite having the wrong reasoning. 
In contrast, GPT-4 stays faithful to its reasoning chain, resulting in a slightly inaccurate final answer.
Note that the overall performance of all models we test is still somewhat low, 
hence not majorly affected by memorization of similar problems.

\begin{table}[ht]
\centering
\caption{Example of memorization in ARB. GPT-3.5-turbo's response indicates memorization of the correct expression.
GPT-4 commits to its reasoning better, resulting in the wrong final answer.}
\vspace{2mm}
\label{table:memorization}
\begin{tabularx}{\textwidth}{p{0.11\textwidth}X}
\toprule
\parboxc{Problem} & \parboxc{
A cylindrical capacitor of length $L$ consists of an inner conductor wire of radius $a$, a thin outer conducting shell of radius $b$. The space in between is filled with nonconducting material of dielectric constant $\varepsilon$.

Find the capacitance.
} \\
\midrule
\parboxc{Reference Solution} & \parboxc{
Supposing that the charge per unit length of the inner wire is $-\lambda$ and using cylindrical coordinates $(r, \varphi, z)$, we find the electric field intensity in the capacitor by Gauss' theorem to be

$$
\mathbf{E}=-\frac{\lambda}{2 \pi \varepsilon r} \mathbf{e}_{r}=\frac{-Q}{2 \pi \varepsilon L r} \mathbf{e}_{r} .
$$The potential difference between the inner and outer capacitors is

$$
V=-\int_{a}^{b} \mathrm{E} \cdot d \mathrm{r}=\frac{\lambda}{2 \pi \varepsilon} \ln \left(\frac{b}{a}\right) .
$$

Hence the capacitance is

$$
C=\frac{\lambda L}{V}=\frac{2 \pi \varepsilon L}{\ln \left(\frac{b}{a}\right)}
$$
} \\
\midrule
\parboxc{\texttt{gpt-3.5} \texttt{-turbo} Solution} & \parboxc{
The capacitance of a cylindrical capacitor is given by the equation:

$$C=\frac{2\pi\varepsilon L}{\ln(b/a)}$$

where $\varepsilon$ is the dielectric constant, $L$ is the length of the capacitor, 
$a$ is the radius of the inner conductor, and $b$ is the radius of the outer conductor.

Therefore, the capacitance of the given cylindrical capacitor is:

$$C=\frac{2\pi\varepsilon L}{\ln(b/a)}=\frac{2\pi\varepsilon L}{\ln(b)-\ln(a)}$$

ANSWER: $\frac{2\pi\varepsilon L}{\ln(b)-\ln(a)}$
} \\
\bottomrule
\end{tabularx}
\end{table}

\begin{table}[ht]
\centering
\caption*{Table \ref{table:memorization} (continued)}
\begin{tabularx}{\textwidth}{p{0.11\textwidth}X}
    \toprule
\parboxc{\texttt{gpt-4} Solution} & \parboxc{
To find the capacitance of the cylindrical capacitor, we will use the general formula for capacitance: 
$C = \frac{Q}{V}$, where $Q$ is the charge stored in the capacitor and $V$ is the potential difference between the inner conductor wire and the outer conducting shell. 
First, let's find the electric field $E$ between the two conductors. 
We can use Gauss's law, which states that the electric flux through a closed surface is equal to the charge enclosed by the surface 
divided by the permittivity of free space $\varepsilon_0$: 
$$\oint \vec{E} \cdot d\vec{A} = \frac{Q_{enclosed}}{\varepsilon_0}.$$ 
We will choose a cylindrical Gaussian surface of radius $r$ and length $L$, where $a < r < b$. 
The electric field is radial, so the electric flux through the curved surface of the cylinder is: 
$$\oint \vec{E} \cdot d\vec{A} = E \oint dA = E (2\pi r L).$$ 
The charge enclosed by the Gaussian surface is the charge on the inner conductor wire, 
which is given by its linear charge density $\lambda$ multiplied by the length of the Gaussian surface, 
$L$: $Q_{enclosed} = \lambda L$. 

Now we can use Gauss's law to find the electric field: 
$$E(2\pi r L) = \frac{\lambda L}{\varepsilon_0} \Rightarrow E = \frac{\lambda}{2\pi r \varepsilon_0}.$$
However, we need to account for the dielectric material in between the conductors. 
The electric field in a dielectric material is related to the electric field in vacuum by the dielectric constant $\varepsilon$: 
$$E_{dielectric} = \frac{E}{\varepsilon}.$$ 

So, the electric field in the dielectric material is: 
$$E_{dielectric} = \frac{\lambda}{2\pi r \varepsilon_0 \varepsilon}.$$ 
Now we can find the potential difference between the inner conductor wire and the outer conducting shell by integrating the electric field along the radial direction: 
$$V = -\int_{a}^{b} E_{dielectric} dr = -\int_{a}^{b} \frac{\lambda}{2\pi r \varepsilon_0 \varepsilon} dr.$$

Integrating, we get:

$$V = -\frac{\lambda}{2\pi \varepsilon_0 \varepsilon} \ln(\frac{b}{a}).$$

Finally, we can find the capacitance using the formula $C = \frac{Q}{V}$:

$$C = \frac{\lambda L}{-\frac{\lambda}{2\pi \varepsilon_0 \varepsilon} \ln(\frac{b}{a})}.$$

Simplifying, we get:

ANSWER: $C = \frac{2\pi \varepsilon_0 \varepsilon L}{\ln(\frac{b}{a})}$.
} \\
\bottomrule
\end{tabularx}
\end{table}
  \providecommand{\parboxc}[1]{\parbox[c]{\hsize}{\vspace{2mm}#1\vspace{2mm}}}

\section{Prompts for \Cref{sec:evaluation,sec:rubric-eval}}
For OpenAI chat models (\texttt{gpt-4} and \texttt{gpt-3.5-turbo}), following best practices, we prepend the system prompt.
We do not use the system prompt for the other models due to there not being a well-established way to do so, and out of concerns of hurting performance on the other models.
We note that omitting the system prompt should have little effect on the performance of the other models, as the directions of the system prompt are essentially restated in our user prompts.

\begin{table}[ht]
\centering
\caption{Prompt used for multiple-choice MCAT and Law problems.}
\vspace{2mm}
\label{prompt:mcat}
\begin{tabularx}{\textwidth}{p{0.11\textwidth}X}
\toprule
\parboxc{System} & \parboxc{
You are a top graduate student taking an open-ended qualifying exam. Your final answer should always be in the last line of your response, preceded by ANSWER:.
} \\
\midrule
\parboxc{User} & \parboxc{
You are a top graduate student taking a qualifying exam. Below you will find a multiple choice question.
\\

Question:
\emph{\{Problem\_Statement\}}
\\

Answer Choices:
\emph{\{Answer\_Choices\}}
\\

Now it is time to choose an answer. 
Think carefully and go step by step. \\ 
Make sure to justify all your work. 
Your final answer should be one of A,B,C,D,... given at the end of your work and preceded by ANSWER:. 
For example, if you think the answer is B, the last line of your answer should be ANSWER: B \\

Solution: 
} \\
\bottomrule
\end{tabularx}
\end{table}

\begin{table}[ht]
\centering
\caption{Prompt used for numerical problems.}
\vspace{2mm}
\label{prompt:numerical}
\begin{tabularx}{\textwidth}{p{0.11\textwidth}X}
\toprule
\parboxc{System} & \parboxc{
You are a top graduate student taking an open-ended qualifying exam. Your final answer should always be in the last line of your response, preceded by ANSWER:.
} \\
\midrule
\parboxc{User} & \parboxc{
You are a top graduate student taking an open-ended qualifying exam. \
Below you will find a question requiring you to compute a numerical value.
\\

Question:
\emph{\{Problem\_Statement\}}
\\

Now it is time to give your answer. Think carefully and go step by step. Make sure to justify all your work. Please simplify all expressions as much as possible and do not leave any variables in your final answer.\\
Your final answer should NOT contain units and should be given at the end of your work and preceded by ANSWER:\\
For example, if you think the answer is $2.4$ meters, the last line of your answer should be ANSWER: $2.4$. \\

Solution: 
} \\
\bottomrule
\end{tabularx}
\end{table}

\begin{table}[ht]
\centering
\caption{Prompt used for symbolic problems.}
\vspace{2mm}
\label{prompt:symbolic}
\begin{tabularx}{\textwidth}{p{0.11\textwidth}X}
\toprule
\parboxc{System} & \parboxc{
You are a top graduate student taking an open-ended qualifying exam. Your final answer should always be in the last line of your response, preceded by ANSWER:.
} \\
\midrule
\parboxc{User} & \parboxc{
You are a top graduate student taking an open-ended qualifying exam. \
Below you will find a question requiring you to give a symbolic answer.
\\

Question:
\emph{\{Problem\_Statement\}}
\\

Now it is time to give your answer. Think carefully and go step by step. Make sure to justify all your work. \\
Your final answer should NOT contain units and should be given at the end of your work and preceded by ANSWER: \\
For example, if you think the answer is $x*y$, the last line of your answer should be ANSWER: $x*y$ \\

Solution: 
} \\
\bottomrule
\end{tabularx}
\end{table}

\begin{table}[ht]
\centering
\caption{Prompt used for proof-like problems.}
\vspace{2mm}
\label{prompt:proof-like}
\begin{tabularx}{\textwidth}{p{0.11\textwidth}X}
\toprule
\parboxc{System} & \parboxc{
You are a top graduate student taking an open-ended qualifying exam. Your final answer should always be in the last line of your response, preceded by ANSWER:.
} \\
\midrule
\parboxc{User} & \parboxc{
You are a top graduate student taking an open-ended qualifying exam. \
Below you will find a question requiring you to prove the given statement.
\\

Question:
\emph{\{Problem\_Statement\}}
\\

Now it is time to give your answer. Think carefully and go step by step. Make sure to justify all your work. \\

Solution: 
} \\
\bottomrule
\end{tabularx}
\end{table}

\begin{table}[ht]
\centering
\caption{Prompt used for GPT-3.5-turbo symbolic evaluation.}
\vspace{2mm}
\label{prompt:symb-eval}
\begin{tabularx}{\textwidth}{p{0.11\textwidth}X}
\toprule
\parboxc{System} & \parboxc{
You are a top professor grading an open-ended qualifying exam.
} \\
\midrule
\parboxc{User} & \parboxc{
 Problem Statement: Give a solution to the differential equation $y'' = -y$\\
Reference Answer: $y(t) = cos(t)$ or $y(t) = sin(t)$\\
Model Answer: $y(x) = sin(x)$\\
Score: The correct answer is either $y(t) = cos(t)$ or $y(t) = sin(t)$. The model gave $y(x) = sin(x)$. Since the function variable was not specified, the model matches one of the reference answers. GRADE: 1 \\

Problem Statement: 
\emph{\{Problem\_Statement\}} \\
Reference Answer:
\emph{\{Final\_Answer\}} \\
Model Answer:
\emph{\{model\_final\_answer\}} \\ 

Now it is time to grade the model answer. \
If the solution is incorrect give GRADE: 0. \
If the solution is nearly correct up to a constant give GRADE: 0.5. \
If the solution is correct give GRADE: 1. \
Before coming to a final grade think think carefully and go step by step. DO NOT TRY TO SOLVE THE PROBLEM. \
If a variable name is not specified and the reference answer and the model answer are the same up to the name of a variable give a score of GRADE: 1. \
For example if the reference answer is $$f(x) = x^2$$ and the model answer is $$f(y) = y^2$$ give a score of GRADE: 1.\\

Score:
    The reference answer 
} \\
\bottomrule
\end{tabularx}
\end{table}

\begin{table}[ht]
\centering
\caption{Prompt used for GPT-4 rubric self-evaluation.}
\vspace{2mm}
\label{prompt:rubric-eval}
\begin{tabularx}{\textwidth}{p{0.11\textwidth}X}
\toprule
\parboxc{System} & \parboxc{
You are a top professor grading an open-ended qualifying exam.
} \\
\midrule
\parboxc{User} & \parboxc{
Problem Statement:
    \emph{\{Problem\_Statement\}} \\
Rubric:
    \emph{\{rubric\}} \\
Student Answer:
    \emph{\{response\}} \\

Now it is time to grade the student answer. Make sure to check each point of the rubric step by step. 
And make sure to print the total number of earned points at the end of your grading. For example, if the student earned 8 points, print Rubric Score: 8 points \\

Rubric Evaluation: 
} \\
\bottomrule
\end{tabularx}
\end{table}

\begin{table}[ht]
\centering
\caption{Prompt used for GPT-4 rubric design.}
\vspace{2mm}
\label{prompt:rubric-design}
\begin{tabularx}{\textwidth}{p{0.11\textwidth}X}
\toprule
\parboxc{System} & \parboxc{
You are a top professor grading an open-ended qualifying exam.
} \\
\midrule
\parboxc{User} & \parboxc{
    Problem Statement:
        \emph{\{Sample\_Problem\_Statement\}}\\
    Reference Solution:
        \emph{\{Sample\_Reference\_Solution\}}\\
    Rubric: 
        \emph{\{Handcrafted\_Rubric\}}\\

Problem Statement: 
    \emph{\{ Problem\_Statement \}}\\
Reference Solution: 
    \emph{\{ Solution \}}\\

Now it is time to write the rubric. Make sure to think carefully and go step by step, \
breaking down the problem into multiple parts. The total number of possible points should sum to 10. \\

Rubric: 
} \\
\bottomrule
\end{tabularx}
\end{table}  \section{Checklist Information.}

\textbf{Dataset Intended Uses.} The ARB benchmark dataset, documented within the paper, is primarily intended for 
research purposes.
We do not advocate for this dataset to train models that help students cheat on exams. 
We hope that the research community will use this benchmark to better assess reasoning capabilities of language models.\\

\textbf{Author Statement and License.} We bear all responsibility in case of violation of rights. 
The ARB dataset is licensed under CC BY 4.0, and all helper code we release is under the MIT license. 
For all problems originating in books listed in \Cref{sec:benchmark-description}, following \citep{MATH},
we abide by Fair Use §107: 
"the fair use of a copyrighted work, including such use by ... scholarship, or research, 
is not an infringement of copyright", where fair use is determined by "the purpose and character of the use, including whether such use is of a commercial nature or is for nonprofit educational purposes" 
and "the effect of the use upon the potential market for or value of the copyrighted work".   \end{appendices}

\end{document}